\pdfoutput=1

\documentclass[11pt]{article}

\usepackage{ACL2023}

\usepackage{times}
\usepackage{latexsym}

\usepackage[T1]{fontenc}

\usepackage[utf8]{inputenc}

\usepackage{microtype}

\usepackage{inconsolata}

\usepackage{array}
\usepackage{multirow}
\usepackage{makecell}
\usepackage{soul}
\usepackage{url}
\usepackage{graphicx}
\usepackage{amsmath}
\usepackage{booktabs}
\usepackage{subfigure}
\usepackage{algorithmic}
\usepackage{enumitem}
\usepackage{amsthm,amsmath,amssymb}
\usepackage{mathrsfs}
\usepackage{bm}
\usepackage{hyperref}
\usepackage[ruled,linesnumbered]{algorithm2e}

\title{EM Pre-training for Multi-party Dialogue Response Generation}

\author{Yiyang Li$^{1,2}$ \and Hai Zhao$^{1,2,}$\thanks{\; Corresponding author. This paper was partially supported by Key Projects of National Natural Science Foundation of China (U1836222 and 61733011).}\\
        $^1$ Department of Computer Science and Engineering, Shanghai Jiao Tong University\\
        $^2$ Key Laboratory of Shanghai Education Commission for Intelligent Interaction\\and Cognitive Engineering, Shanghai Jiao Tong University\\
        \texttt{eric-lee@sjtu.edu.cn, zhaohai@cs.sjtu.edu.cn}
}

\begin{document}
\maketitle

\begin{abstract}
Dialogue response generation requires an agent to generate a response according to the current dialogue history, in terms of which two-party dialogues have been well studied, but leaving a great gap for multi-party dialogues at the same time. Different from two-party dialogues where each response is a direct reply to its previous utterance, the addressee of a response utterance should be specified before it is generated in the multi-party scenario. Thanks to the huge amount of two-party conversational data, various pre-trained language models for two-party dialogue response generation have been proposed. However, due to the lack of annotated addressee labels in multi-party dialogue datasets, it is hard to use them to pre-train a response generation model for multi-party dialogues. To tackle this obstacle, we propose an Expectation-Maximization (EM) approach that iteratively performs the expectation steps to generate addressee labels, and the maximization steps to optimize a response generation model. Theoretical analyses and extensive experiments have justified the feasibility and effectiveness of our proposed method. The official implementation of this paper is available at \url{https://github.com/EricLee8/MPDRG}.
\end{abstract}

\section{Introduction}
Inspired by the tremendous success in pre-training large language models (PLMs) in general domains \cite{devlin2019bert, clark2020electra, GPT}, efforts have been made to train PLMs for dialogue response generation \cite{zhang-etal-2020-dialogpt, bao-etal-2020-plato, chen-etal-2022-dialogved}. However, they constrain the dialogues to be either two-party, or sequential structured (i.e. each utterance replies directly to its previous utterance). Different from them, a multi-party dialogue can involve multiple interlocutors, where each interlocutor can reply to any preceding utterances, making the response relations of the dialogue tree-structured and much more complicated \cite{zhang_aaai2018_addr, le-etal-2019-whoisspeaking, minlie, wang-etal-2020-response-in-multi-party}. Besides, the speaker and addressee of a response utterance should be specified before it is generated in multi-party scenario, making the annotated data for multi-party dialogue response generation (MPDRG) less available.

\begin{figure}[tbp]
    \centering
    \includegraphics[width=0.46\textwidth]{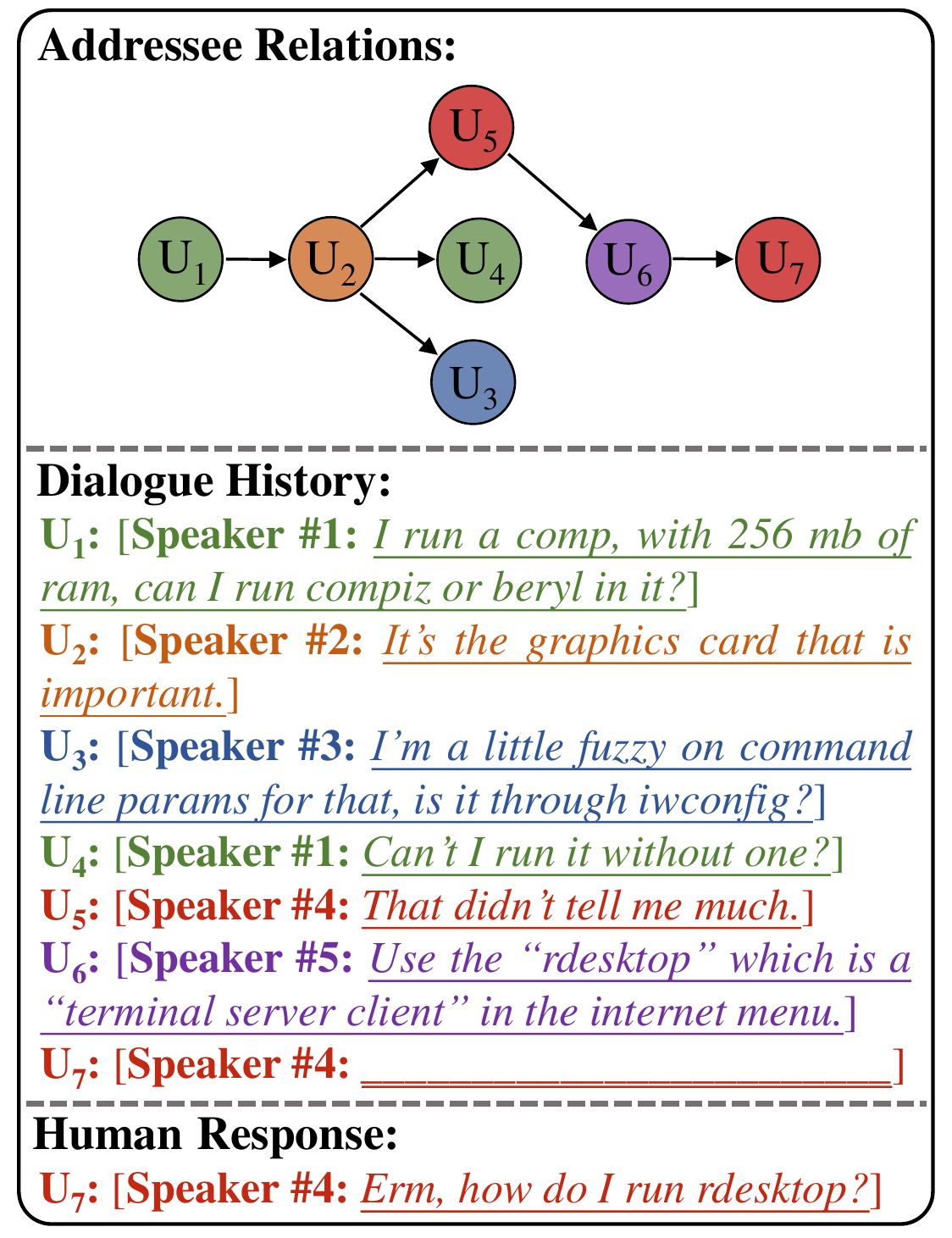}
    \caption{An example of multi-party dialogue response generation task, better view in color.}
    \label{fig:multi_example}
\end{figure}

Figure \ref{fig:multi_example} illustrates an example of MPDRG task taken from the Ubuntu IRC benchmark \cite{GSN}. The upper part shows the tree-structured addressee relations of the dialogue, where the arrows point from addressees to speakers, and different colors represent different interlocutors. The middle part displays the content of the dialogue history, where $\operatorname{U_7}$ is the response to be generated. The addressee ($\operatorname{U_6}$) and the speaker (\#4) of it are given, and the content of this response is the target of our model. The lower part gives the human response, which is also called the ground truth reference.

Previous works on MPDRG fine-tune generative PLMs on small multi-party dialogue datasets with explicit addressee annotations. They utilize the response annotations to form a tree-structured response graph, then encode the dialogue history using either homogeneous or heterogeneous Graph Neural Networks (GNNs) \cite{GSN, gu-etal-2022-hetermpc}. Nevertheless, none of them make attempts to pre-train a response generation model for multi-party dialogues due to the lack of large-scale corpora with annotated addressee labels.

To solve the aforementioned problem of data scarcity, we propose an EM approach that iteratively performs the expectation steps to generate addressee labels, and the maximization steps to optimize a response generation model. Specifically, we treat the addressee of each utterance in the dialogue history as a discrete latent variable $z$. During the E-steps, given the current dialogue history $c_t$ and the the response utterance $r_t$, we model the distribution of the current addressee $z_t$ as $p(z_t|c_t,r_t;\bm{\theta})$, where $\bm{\theta}$ is the current model parameters. During the M-steps, we sample $(c_t, r_t, z_t)$ triplets from distribution $p(z_t|c_t,r_t;\bm{\theta})$ and optimize the generative model $p(r_t|c_t, z_t; \bm{\theta})$ on these samples. With the iteration number increasing, the accuracy of latent variable prediction and the quality of generated responses will grow together. It is worth noting that during these iterations, annotated addressee labels are not required, which makes it possible to leverage the huge amount of multi-party dialogue corpora without addressee labels. We provide theoretical analyses to prove the feasibility of our EM method, and conduct experiments on the Ubuntu IRC benchmark, which is used in previous works \cite{GSN, gu-etal-2022-hetermpc}.

The contributions of our work can be summarized as the following three folds:
\begin{itemize}[leftmargin=*, topsep=1pt]
    \setlength{\itemsep}{0pt}
    \setlength{\parsep}{0pt}
    \setlength{\parskip}{0pt}
    \item To the best of our knowledge, we are the first to study the pre-training of multi-party dialogue response generation, which is much more challenging and complicated than two-party dialogues.
    \item We put forward an EM approach to alleviate the scarcity of multi-party dialogue data with addressee labels, making it possible to pre-train a model with huge amount of unlabeled corpora.
    \item We provide theoretical analyses to prove the feasibility of our EM pre-training method, and experimental results on the Ubuntu IRC benchmark show our pre-trained model achieves state-of-the-art performance compared with previous works.
\end{itemize}

\section{Related Works}
\subsection{Pre-training for Response Generation}
\label{sec:related1}
In recent years, researchers have gradually drawn their attention from retrieval-based dialogue systems to generation-based ones. Thanks to the huge amount of two-party dialogue corpora, various PLMs for two-party dialogue response generation have been proposed.

\citet{zhang-etal-2020-dialogpt} propose DialoGPT, which utilizes the sequential response chains in the Reddit Corpus to pre-train an auto-regressive response generation model based on the architecture of GPT \cite{GPT}. Different from their work, which focuses on sequential dialogue history, our work aims to solve the case where the agent can respond to any previous utterance in a tree-structured dialogue history.

\citet{bao-etal-2020-plato} propose PLATO, which models the conversational intents as $K$ discrete latent variables, then utilizes response selection, bag-of-words prediction, and language modeling objectives to train the model. DialogVED \cite{chen-etal-2022-dialogved} further extends the discrete latent variables to continuous ones, and models them with a multi-variable Gaussian distribution. It utilizes KL divergence reduction to optimize the parameters of the latent distribution and applies masked language modeling, response generation, and bag-of-words prediction to train the whole model. PLATO and DialogVED focus on two-party conversations, and the conversational intents they put forward have no corresponding concepts of actual entities (e.g., intent to argue, intent to end a conversation, and so on). Distinct from their works, we lay emphasis on multi-party dialogues, and the latent variables of our method have actual meanings: variable $z_t = j$ indicates that the addressee of the response at the $t_{th}$ turn is the $j_{th}$ utterance.

\begin{figure*}[tbp]
    \centering
    \includegraphics[width=0.93\textwidth]{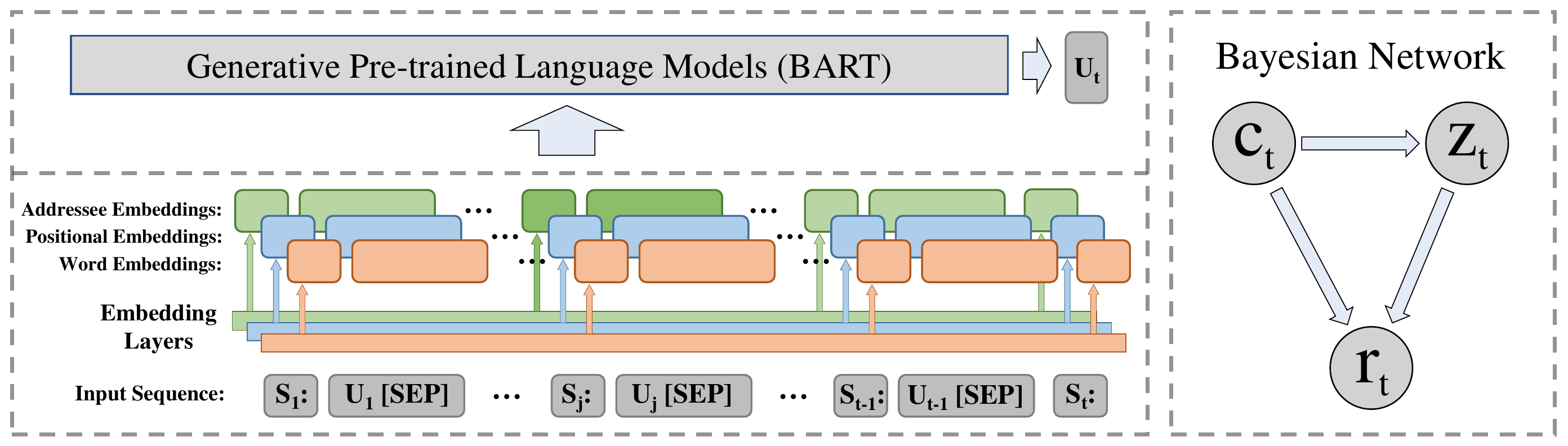}
    \caption{The overview of our model architecture. The left part shows how we incorporate the addressee information into response generation by adding addressee embeddings. The right part illustrates a Bayesian Network of how a response is generated given the current dialogue history $c_t$ and the addressee $z_t$.}
    \label{fig:model_overview}
\end{figure*}

\subsection{Multi-party Dialog Response Generation}
\label{sec:related2}
Several previous works have studied the MPDRG task. \citet{GSN} extract a subset of the Ubuntu Dialogue Corpus \cite{lowe-etal-2015-ubuntu} with explicit addressee labels to construct the Ubuntu IRC benchmark, where they propose a Graph Structured Neural Network (GSN) for dialogue modeling. Specifically, they first treat each utterance of a dialogue as a node, and the addressee relations as edges to construct a dialogue graph, then make use of GNNs to encode the dialogue history. Finally, they adopt a Gated Recurrent Unit (GRU) with cross attention as the decoder to generate responses. \citet{gu-etal-2022-hetermpc} put forward HeterMPC, which models the dialogue history as a heterogeneous graph. In detail, they first design six types of edges: reply and replied-by, address and addressed-by, speak and spoken-by, among two kinds of nodes: interlocutor nodes and utterance nodes, and then encode the dialogue history using Transformers \cite{attention} together with heterogeneous GNNs. Finally, they utilize a Transformer Decoder to generate responses. Instead of fine-tuning models on a small dataset with annotated addressee labels as these existing work did, our work focuses on the utilization of large unlabeled corpora to pre-train a response generation model for multi-party dialogues.

\section{Methodology}
To design a model for multi-party dialogue response generation and make it compatible with the EM training algorithm, there are two important things to consider: how to model $p(r_t|c_t, z_t; \bm{\theta})$ in the maximization step, and how to compute $p(z_t|c_t, r_t;\bm{\theta})$ in the expectation step. In this section, we will first address these two problems, then mathematically derive the feasibility of our EM pre-training algorithm.

\subsection{Task Formulation}
Given an input sequence of the dialogue history and the speaker of the response at time step $t$, $\mathbb{X} = \{\operatorname{S_1: U_1 [SEP] S_2: U_2 [SEP] \dots S_{t-1}: U_{t-1} [SEP] S_t:}\}$, together with the addressee of the response $z_t = j$, our goal is to train a model that can generate an response $\mathbb{Y} = \operatorname{U_t}$. Here each $\operatorname{S_i}$ is the name of the speaker at time step $i$, which is represented as \emph{Speaker \#$S_i$} like those in Figure \ref{fig:multi_example}. $\operatorname{U_i} = \{w_{i1}, w_{i2},\dots, w_{in_i}\}$ is the content of the $i_{th}$ utterance with $n_i$ words. $z_t = j$ represents that $\operatorname{S_t}$ speaks to $\operatorname{S_j}$, who utters $\operatorname{U_j}$, and $\operatorname{[SEP]}$ is a special token that indicates the end of a dialogue turn.

\subsection{Addressee Modeling}
\label{sec:addr_modeling}
In this section, we answer the first question: how to model $p(r_t|c_t, z_t; \bm{\theta})$, or in other words, how to incorporate the addressee information $z_t = j$ into the process of generating a response $r_t$. We design a straightforward method that adds addressee embeddings to the positional encodings and word embeddings, before they are further encoded by a PLM. The left part of Figure \ref{fig:model_overview} illustrates this method, where we use an embedding look-up table with $2$ entries to indicate whether a word belongs to the addressee utterance or not. Specifically, if a word is in the addressee utterance, it will get its addressee embedding from entry $1$, otherwise from entry $0$. Since addressee modeling is not the key contribution of this work, we just adopt the most straightforward and effective way. In our experiments, we use BART \cite{lewis-etal-2020-bart} as the backbone PLM, following previous works \cite{gu-etal-2022-hetermpc}. Due to the page limit, the proverbial architecture of Transformer and BART are omitted here.

\subsection{Latent Variable Prediction}
\label{sec:bayesian}
In this section, we answer the second question: how to compute $p(z_t|c_t, r_t; \bm{\theta})$ in the expectation step, or in other words, how to predict the distribution of the unlabeled addressee $z_t$, given the current dialogue context $c_t$, response $r_t$, under parameters $\bm{\theta}$. The solution to this question is essentially the most important part of our method since it delicately solves the problem of data scarcity in MPDRG.

Let's consider what humans will do to participate in a multi-party conversation. First, we will read the dialogue history $c_t$, then choose an addressee $z_t$ to reply. Once $c_t$ and $z_t$ are determined, we will utter a response according to the content of the whole dialogue and the addressee utterance. The right part of Figure \ref{fig:model_overview} gives the Bayesian Network of the above process, where the joint distribution of $(c_t, z_t, r_t)$ can be factorized as:
\begin{equation}
    \label{eq:bayesian}
    p(c, z, r) = p(c)\cdot p(z|c) \cdot p(r|c, z)
\end{equation}
Here we omit the subscript $t$ and model parameters $\bm{\theta}$ for simplicity. Given Eq. (\ref{eq:bayesian}), $p(z|c, r; \bm{\theta})$ can be derived as:
\begin{equation}
\label{eq:prior}
    \begin{aligned}
        p(z|c,r) &= \frac{p(c, z, r)}{p(c, r)}\\
        &= \frac{p(c)\cdot p(z|c)\cdot p(r|c, z)}{p(c)\cdot p(r|c)}\\
        &= \frac{p(z|c)\cdot p(r|c, z)}{p(r|c)}
    \end{aligned}
\end{equation}
We assume that the probability of choosing any previous utterance as the addressee is the same given the current dialogue history, which means $p(z|c)$ obeys a uniform distribution. Meanwhile, the denominator $p(r|c)$ is independent of $z$, leaving only the term $p(r|c, z)$. Now, we can induce that:
\begin{equation}
    \label{eq:unnormal}
    p(z|c, r) \propto p(r|c, z)
\end{equation}
Therefore, for each $z^i, i=1, 2,\dots, t-1$, we have:
\begin{equation}
    \label{eq:e_step}
    p(z^i|c, r) = \frac{p(r|c, z^i)}{\sum_{j=1}^{t-1} p(r|c, z^j)}
\end{equation}
In practice, we can use the generative model $p(r_t|c_t, z_t; \bm{\theta})$ to compute the probability distribution of $p(z_t|c_t, r_t; \bm{\theta})$ by Eq. (\ref{eq:e_step}).

\begin{figure}[tbp]
    \centering
    \includegraphics[width=0.48\textwidth]{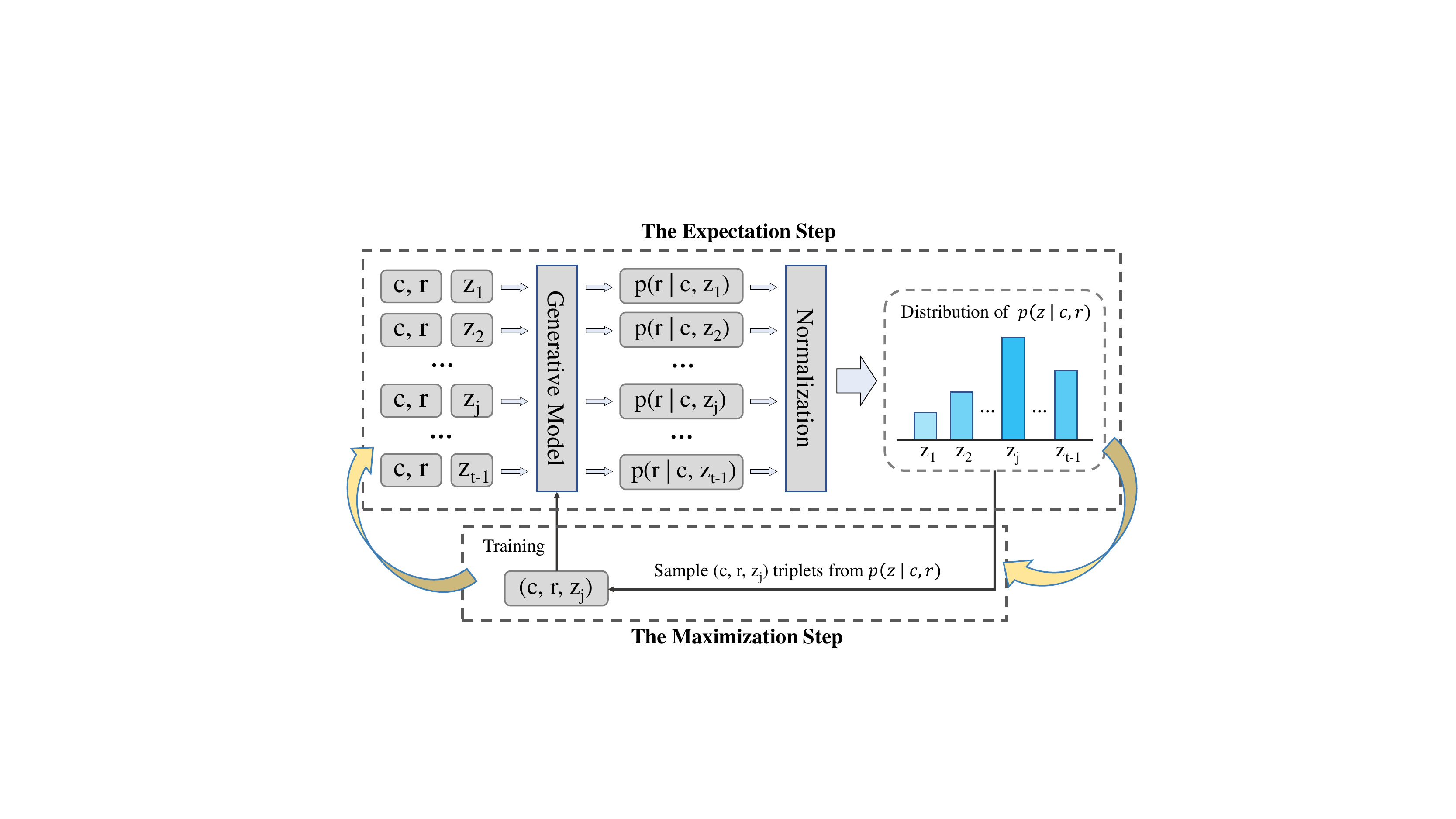}
    \caption{The overview of the EM process, where the expectation steps and maximization steps are performed alternately and iteratively.}
    \label{fig:em}
\end{figure}

\subsection{Expectation-Maximization Process}
\label{sec:EM}
Figure \ref{fig:em} illustrates the overview of our EM training process. During the E-steps, we compute the probability distribution of the latent variable (the addressee $z$). During the M-steps, we sample $(c, r, z)$ triplets from this distribution and optimize the generative model by standard training algorithms.\\
\textbf{The Expectation Step} is to compute the conditional distribution of the latent variable $z_t$, given the observed data $(c_t, r_t)$ and the current model parameters $\bm{\theta}$, where Eq. (\ref{eq:e_step}) gives a reasonable approximation of this value. Specifically, for a sample $(c_t, r_t)$, with the model parameters $\bm{\theta}$ fixed, we first calculate the un-normalized probability of each of the $i_{th}$ ($i<t$) utterance being the addressee: $p(r_t|c_t, z^i_t; \bm{\theta})$ using Eq. (\ref{eq:unnormal}), then normalize them to get the conditional distribution of $z_t$ using Eq. (\ref{eq:e_step}). Once $P(z_t|c_t, r_t; \bm{\theta})$ is obtained, we sample $(c_t, r_t, z_t)$ triplets from this distribution, which is further used in the maximization step.\\
\textbf{The Maximization Step} is analogical to the normal training process. Given the sampled $\{(c_t^k, r_t^k, z_t^k)\}_{k=1}^N$ triplets, where $N$ is the total number of samples, our goal is to minimize the auto-regressive language modeling loss:
\begin{equation}
    \mathcal{L}_G= -\sum_{k=1}^{N}\sum_{i=1}^{n_k} \log p\left(w^k_i \mid w^k_{<i}, c^k_t, z^k_t; \bm{\theta}\right)
\end{equation}
where $w^k_i$ is the $i_{th}$ word in the response of the $k_{th}$ sample: $r^k_t = \{w^k_i\}_{i=1}^{n_i}$, and $n_i$ is the length of this response.\\
\textbf{Compared with the vanilla EM algorithm}, there are several differences in our implementations. First of all, we do not use the initial model to generate the training data for the first round of the maximization step. Instead, we utilize the discourse parser provided by \citet{minlie} to predict the addressee of each utterance in the unlabeled corpus to get a coarse initial training dataset. The reason for this initialization method is that the initialization of training data (or model parameters) is vital to the EM method, which helps it converge to a better point. Second, rather than sampling $z_t$ from its conditional distribution, we adopt a hard EM approach which takes the value $z_t^i$ with highest probability as the predicted label, where $i = \mathop{\arg\max}\limits_{i} p(z_t^i|c_t, r_t; \bm{\theta})$. This hard EM approach is proved as more effective to boost the performance \cite{EMQA}. Finally, to ensure the quality of the generated training data in the maximization step, we set a hyper-parameter $\alpha \in [0, 1]$ to control the proportion of training data that is actually used. Specifically, we first rank the prediction confidence of each $z_t^k$ according to the value of $p(z_t^k|c_t^k, r_t^k; \bm{\theta})$, then pick the top $\alpha\times N$ samples with the highest confidence scores. In our experiments, $\alpha$ is dynamically set to ensure the addressee prediction accuracy of the selected samples is over $80\%$ in an annotated validation set.

\subsection{Proof of Feasibility}
\label{sec:proof}
In a multi-party dialogue corpus without annotated addressee labels, a usual solution to train a response generation model is to maximize the marginal log-likelihood (or incomplete log-likelihood) over all possible addressees:
\begin{equation}
    \label{eq:marginal}
    \ell(c, r; \bm{\theta}) = \rm{log}\ p(r|c; \bm{\theta}) = \rm{log} \sum_i p(r, z_i|c; \bm{\theta})
\end{equation}
However, this objective is hard to optimize since the distribution of $z$ is hard to obtain. Here, we define an expected complete log-likelihood where our estimation of $p(z_t|c_t, r_t;\bm{\theta})$ can come to rescue:
\begin{equation}
\label{eq:if_and_only_if}
    \begin{aligned}
        \hat{\ell}(c, r; \bm{\theta}) &= q(z_i)\sum_i \rm{log}\ p(r,z_i|c; \bm{\theta})\\
        q(z) &= p(z_t|c_t, r_t;\bm{\theta})
    \end{aligned}
\end{equation}
Our new objective now becomes maximizing the expected complete log-likelihood. The relation between $\ell$ and $\hat{\ell}$ can be derived as follows:
\begin{equation}
    \begin{aligned}
        \ell(c, r; \bm{\theta}) &= \rm{log} \sum_i p(r, z_i|c; \bm{\theta})\\
        &= \rm{log}\sum_i q(z_i)\cdot \frac{p(r, z_i|c; \bm{\theta})}{q(z_i)}\\
        &\ge \sum_i q(z_i)\cdot \rm{log}\frac{p(r,z_i|c;\bm{\theta})}{q(z_i)}\\
        &= \sum_i q(z_i)\cdot \rm{log}\ p(r,z_i|c;\bm{\theta})\\
        &\quad \quad - \sum_i q(z_i)\cdot \rm{log}\ q(z_i)\\
        &= \hat{\ell}(c, r; \bm{\theta}) + \mathcal{H}_{q(z)}
    \end{aligned}
\end{equation}
where the third line is derived from the \emph{Jensen Inequality}, and $\mathcal{H}_{q(z)}$ is the entropy of the distribution of $z$. Since $\mathcal{H}_{q(z)}\ge 0$, we can derive that $\hat{\ell}(c, r; \bm{\theta}) \leq \ell(c, r; \bm{\theta})$, which means $\hat{\ell}$ is the lower bound of $\ell$. By maximizing the lower bound $\hat{\ell}$, we can indirectly maximize $\ell$, which is originally hard to optimize. Another important observation is hat $\hat{\ell} = \ell$ if and only if $q(z) = p(z_t|c_t, r_t;\bm{\theta})$, which is exactly what we calculate during the E-steps in Eq. (\ref{eq:if_and_only_if}). Though the derivation of the posterior distribution of $z$ is not exact since we assume uniform prior in Eq. (\ref{eq:prior}), it is still much closer to the real distribution compared to random $q(z)$.

It is worth noting that the global optimal point is not guaranteed to be reached by this algorithm, and it depends heavily on the initialization of model parameters or the training data for the first round of the maximization step. This explains the reason why we utilize a discourse parser to get a coarse initial training dataset instead of using the expectation step at the first iteration in Section \ref{sec:EM}.

\begin{table*}[tbp]
    \centering
    \small
    \begin{tabular}{lrrrrrr}
        \specialrule{0.09em}{0.0pt}{1.8pt}
        Model &  BLEU-1 & BLEU-2 & BLEU-3 & BLEU-4 & METEOR & ROUGE-L\\
        \specialrule{0.03em}{1.3pt}{0.3pt}
        \specialrule{0.03em}{0.3pt}{1.3pt}
        GPT-2 \cite{GPT} & $10.37$ & $3.60$ & $1.66$ & $0.93$ & $4.01$ & $9.53$\\
        GSN \cite{GSN} & $10.23$ & $3.57$ & $1.70$ & $0.97$ & $4.10$ & $9.91$\\
        HeterMPC$\operatorname{_{BART}}$ \cite{gu-etal-2022-hetermpc} & $12.26$ & $4.80$ & $2.42$ & $1.49$ & $4.94$ & $11.20$\\
        \specialrule{0.06em}{1.3pt}{1.3pt}
        BART \cite{lewis-etal-2020-bart} & $11.25$ & $4.02$ & $1.78$ & $0.95$ & $4.46$ & $9.90$\\
        \quad Pre-training Only (PO) & $11.78$ & $4.67$ & $2.38$ & $1.41$ & $4.98$ & $11.19$\\
        \quad Fine-tuning Only (FO) & $11.47$ & $5.11$ & $2.98$ & $2.11$ & $5.23$ & $11.31$\\
        \quad Pre-training + Fine-tuning (PF) & $\mathbf{12.31}$ & $\mathbf{5.39}$ & $\mathbf{3.34}$ & $\mathbf{2.45}$ & $\mathbf{5.52}$ & $\mathbf{11.71}$\\
        \specialrule{0.06em}{1.3pt}{1.3pt}
        \quad FO + Reply-Chain & $9.11$ & $3.52$ & $1.99$ & $1.35$ & $4.32$ & $9.36$\\
        \quad PO w/o EM & $10.03$ & $3.90$ & $2.03$ & $1.18$ & $4.56$ & $9.66$\\
        \quad PF w/o EM & $11.39$ & $5.04$ & $3.02$ & $2.15$ & $5.27$ & $11.20$\\
        \quad Denoising + Fine-tuning & $11.49$ & $5.08$ & $3.02$ & $2.13$ & $5.25$ & $11.28$\\
        \specialrule{0.09em}{1.2pt}{0.0pt}
    \end{tabular}
    \caption{Results on the Ubuntu IRC benchmark, where the upper part presents models of previous works, the middle part shows our backbone model BART together with our method under different settings, and the lower part shows the ablation studies.}
    \label{tab:results}
\end{table*}

\section{Experiments}
In this section, we first introduce the datasets to pre-train and evaluate our model, then present the experimental results and comparisons with previous methods.

\subsection{Datasets and Experimental Setups}
For pre-training, we adopt the second version of Ubuntu Dialogue Corpus \cite{lowe-etal-2015-ubuntu}, which contains no annotated addressee labels. The original dataset contains $1$M dialogues for training, and $0.5$M dialogues for validation and testing, respectively. Dialogues that contain less than $4$ turns, or have overlap with the dataset for the downstream task (the Ubuntu IRC benchmark, \citealt{GSN}), are excluded from the pre-training data. After filtering, we eventually get a pre-training dataset that contains 764,373 dialogues.

For fine-tuning, we follow previous works \cite{GSN, gu-etal-2022-hetermpc} to adopt the Ubuntu IRC benchmark, which is constructed by extracting all utterances with response addressees indicated by the ``@" symbol in the Ubuntu Dialogue Corpus. In total, this dataset consists of 311,725 dialogues for training, and 5,000 dialogues for validation and testing, respectively. It is worth noting that this dataset contains addressee labels for every single utterance in the dialogue history, which are utilized by previous methods, yet not by ours.

For both pre-training and fine-tuning, BART \cite{lewis-etal-2020-bart} is used as the backbone model. Before pre-training, we initialize the pre-trained weights from BART-base. During the process of pre-training, we evaluate our model on the validation set of the Ubuntu IRC benchmark, and the best checkpoint is saved for the fine-tuning process.

\subsection{Baseline Models and Evaluation Metrics}
Table \ref{tab:results} shows the results of our method and previous models, where GPT-2, GSN, and HeterMPC \cite{GPT, GSN, gu-etal-2022-hetermpc} are introduced in section \ref{sec:related1} and \ref{sec:related2}, respectively. BART is a sequence-to-sequence model with encoder-decoder Transformer architecture and is trained using denoising objectives. Following \citet{GSN}, we also adopt BLEU-1 to BLEU-4, METEOR, and ROUGE-L as the automatic evaluation metrics, which can be calculated using the \emph{pycocoevalcap} package. Besides automatic evaluation, human evaluation is also conducted and will be introduced in Section \ref{sec:human_eval}.

\subsection{Automatic Evaluation Results}
Let's firstly focus on the upper and middle part of Table \ref{tab:results}, where we present the results of previous models and our methods. Three settings of our method based on BART are experimented with: pre-training only (PO), fine-tuning only (FO), and pre-training-fine-tuning (PF). Results of PO are obtained by directly using the pre-trained model to generate the response for each dialogue. FO means the checkpoint of BART is directly fine-tuned on the Ubuntu IRC benchmark without pre-training. PF follows a pre-training-fine-tuning paradigm, where the best checkpoint of the pre-training process is further fine-tuned on the downstream dataset.

Three observations can be seen from the table. First of all, solely pre-training with our proposed EM method with unlabeled corpus is already able to achieve comparable results with the previous state-of-the-art (SOTA) models. It is surprising since the pre-training requires no annotated addressee labels, while previous models not merely utilize the addressee information of the response utterance, but also make use of the addressee labels of the dialogue history to form a response graph. Second, fine-tuning our model on the downstream dataset with the ground truth addressee labels yields better results compared with pre-training only. Since it uses the ground truth addressee labels of responses, the results of it can be regarded as an upper bound of what the EM training can achieve. Besides, FO outperforms the previous SOTA model by large margins with even simpler architecture and fewer annotations (without addressee labels in the dialogue history), demonstrating the effectiveness of our proposed addressee embeddings. Finally, by further fine-tuning the pre-trained checkpoint with the ground truth addressee labels, we achieve the best performance on all metrics, which shows the transferability of our pre-trained model.

\begin{table}[tbp]
    \centering
    \small
    \begin{tabular}{lrrr}
        \specialrule{0.09em}{0.0pt}{1.8pt}
        Model & Score & Kappa & Best (\%)\\
        \specialrule{0.03em}{1.3pt}{0.3pt}
        \specialrule{0.03em}{0.3pt}{1.3pt}
        Human References & $2.20$ & $0.56$ & $28.00$\\
        \specialrule{0.06em}{1.3pt}{1.3pt}
        BART & $1.68$ & $0.45$ & $8.00$\\
        HeterMPC$\operatorname{_{BART}}$ & $1.88$ & $0.48$ & $8.00$\\
        Ours (PF) & $\mathbf{1.92}$ & $0.47$ & $\mathbf{28.00}$\\
        \specialrule{0.09em}{1.2pt}{0.0pt}
    \end{tabular}
    \caption{Human evaluation results, where \emph{Score} is the average score and \emph{Best} means the ratio of each system being the best response.}
    \label{tab:human_results}
\end{table}

\subsection{Human Evaluation Results}
\label{sec:human_eval}
For human evaluation, we recruit a team with 8 members who have at least a Bachelor's degree in Computer Science and are familiar with Ubuntu and Linux. We randomly sample $100$ examples from the testing set, then ask the team members to score each prediction and select the best one. The quality scores are considered in terms of three independent aspects: 1) relevance, 2) fluency and 3) informativeness. They are scored from 0-3 and the average values were reported. The evaluation results are shown in Table \ref{tab:human_results}, where our model (Pre-training + Fine-tuning) constantly outperforms vanilla BART and the previous SOTA model HeterMPC$\operatorname{_{BART}}$. We also report the Fleiss’s Kappa to indicate the agreement between annotators. Besides, the ratio of our predictions being the best response is the same as that of human responses, demonstrating the high quality of the generated responses of our model.

\section{Analysis}
In order to get more insights into the proposed EM pre-training method, we dive deeper into it by conducting extensive analyses.

\subsection{Ablation Study}
We conduct ablation studies to investigate the contribution of our different designs, whose results are tabulated in the lower part of Table \ref{tab:results}.

Firstly, let's focus on the first line of the lower part. To study whether other utterances that are not in the reply chain of the current addressee can help to generate a better response, we extract the reply train by traversing from the current leave utterance (the response) up to the root node (the first utterance), then train a model by inputting this chain only. We see a large performance drop on all metrics in this setting, demonstrating the significance of the side information provided by the whole context.

Second, let's pay attention to the second and third lines of the lower part. In order to study the effect of the EM pre-training process, which is the key contribution of our work, we remove this process and pre-train a model using only the addressee labels obtained from the discourse parser (i.e. the initial training data used in the first iteration of our EM approach). A sharp performance drop is observed compared with PO and PF with our proposed EM pre-training strategy, demonstrating the significance of our design. Without the iterative EM procedure, the noisy addressee labels obtained from the discourse parser can cause error propagation, which makes the model learn noisy features to predict a response, and hurts the performance.

Finally, aiming at investigating whether the performance gains come from seeing more in-domain data in the pre-training process, we use the same pre-training data to train another model with the denoising objectives proposed in BART \cite{lewis-etal-2020-bart}, then also fine-tune it on the Ubuntu IRC benchmark. The last line of the lower part presents the results, where we observe nearly the same performance compared with FO. This observation indicates that simply performing domain adaptation using the general pre-training objectives is insufficient to benefit the MPDRG task.

\begin{figure}[tbp]
    \centering
    \includegraphics[width=0.48\textwidth]{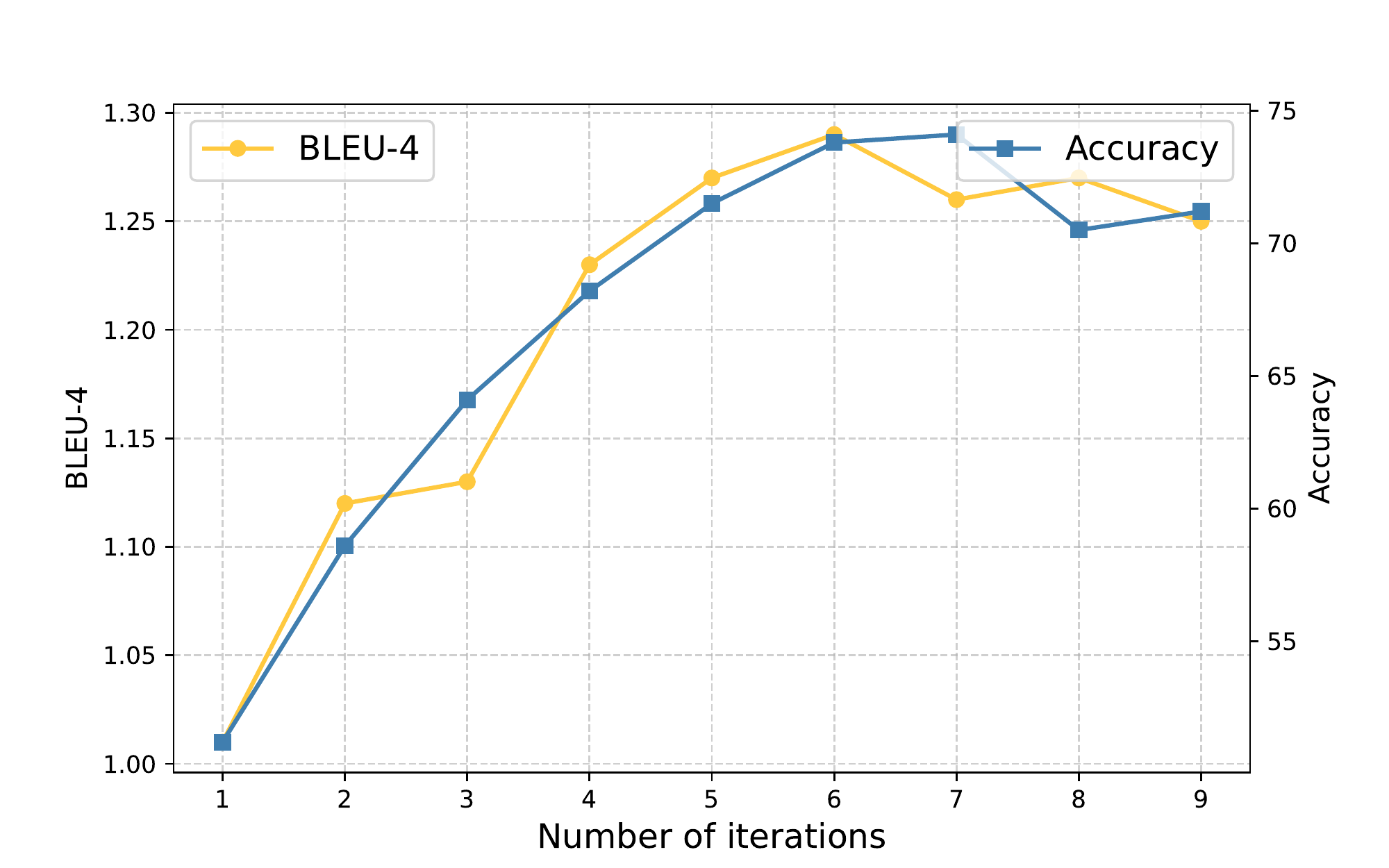}
    \caption{Line chart of the BLEU-4 score and addressee prediction accuracy with the increase of EM iterations.}
    \label{fig:em_iter}
\end{figure}

\subsection{Response Generation vs. Addressee Prediction}
In Section \ref{sec:bayesian}, we prove that $p(z|c, r)\propto p(r|c, z)$. To verify the correctness of this equation and also to investigate the training process of our EM strategy, we draw the line chart of the BLEU-4 score and addressee prediction accuracy of the top-30\% confidence samples on the validation set with the increasing of pre-training iterations. The addressees are predicted using Eq. (\ref{eq:e_step}), where we take the $z^i$ with the highest conditional probability as the predicted addressee.

Figure \ref{fig:em_iter} illustrates the trending of the BLEU-4 score and addressee prediction accuracy. On the one hand, we see that the trending of both metrics is consistent, which means with a more powerful response generation model comes a higher addressee prediction accuracy. This observation verifies the correctness of Eq. (\ref{eq:unnormal}). On the other hand, with the increasing of iterations, both metrics grow mutually, then reach their tops at around the $6_{th}$ iteration, demonstrating the effectiveness of the EM process.

\subsection{Case Studies}
To understand the effect of our method intuitively, we sample two cases from the testing set and present them in this section.

\begin{figure}[tbp]
    \centering
    \includegraphics[width=0.48\textwidth]{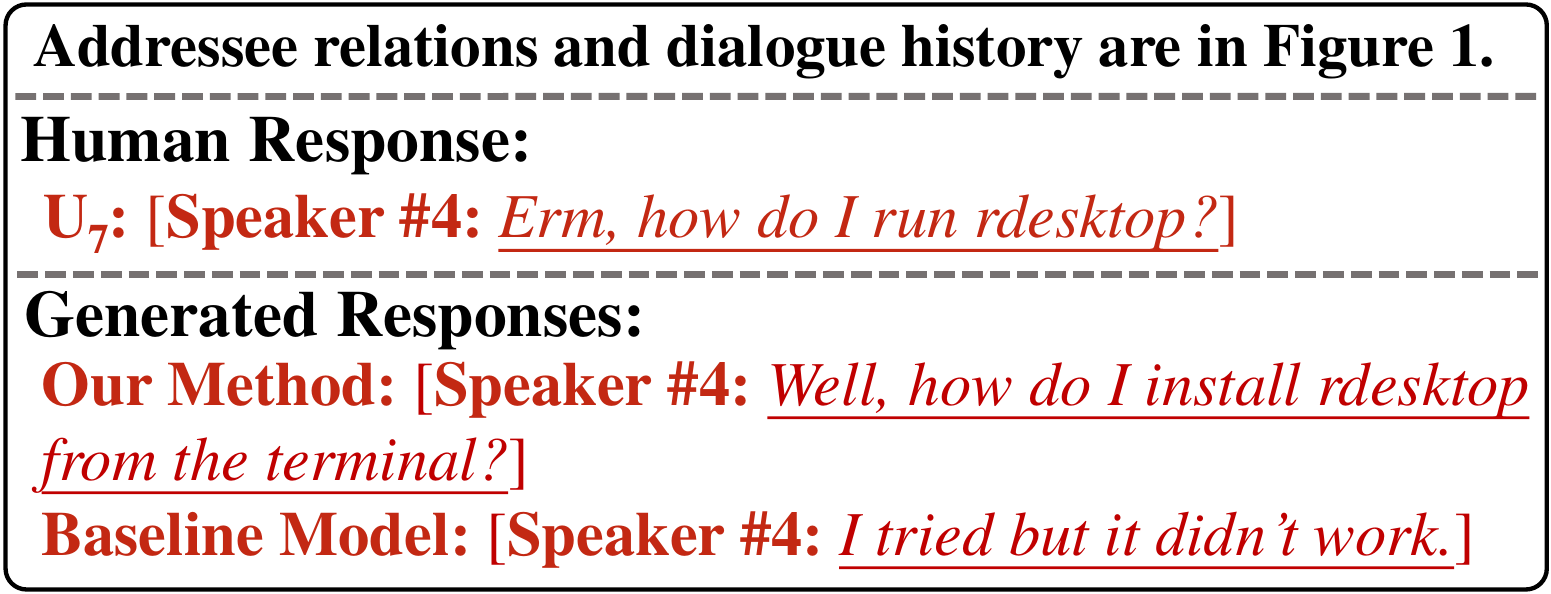}
    \caption{The first example of Case Studies, which shows the generated responses of our model and the baseline model.}
    \label{fig:case2-1}
\end{figure}

Figure \ref{fig:case2-1} illustrates an example whose addressee relations and dialogue history are shown in Figure \ref{fig:multi_example}. This conversation is about how to run the \emph{compiz} or \emph{beryl} in a \emph{comp} with $256$MB RAM. \emph{Speaker \#2} points that \emph{it's the graphic card that is important}, but \emph{Speaker \#4} seems unsatisfied by saying \emph{that didn't tell me much}. After that, \emph{Speaker \#5} suggests using the \emph{rdesktop} and \emph{Speaker \#4} replies him/her. Our model is able to capture the key information \emph{rdesktop} and \emph{terminal} in the addressee utterance $\bm{U_6}$, and generate a proper response \emph{Well, how do I install rdesktop from the terminal}, which is very close to the human answer and even better with more information \emph{from the terminal}. On the contrary, the baseline model (BART) fails to capture the addressee information and just replies with a safe response \emph{I tried but it didn't work}. This case shows the great significance of modeling the addressee information, and also demonstrates the effectiveness of our model design.

\begin{figure}[tbp]
    \centering
    \includegraphics[width=0.45\textwidth]{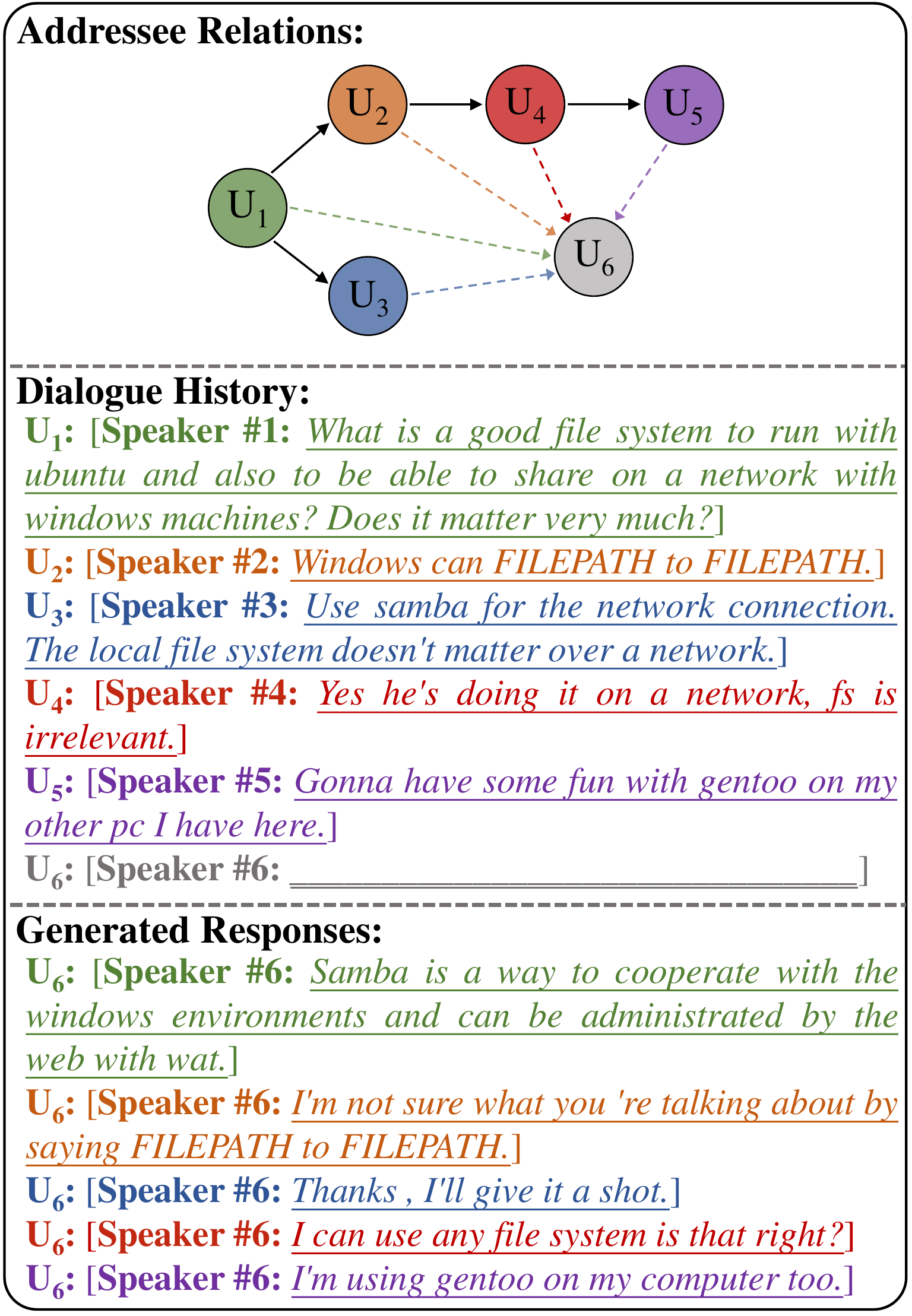}
    \caption{The second example of Case Studies, which illustrates the generated response of our model given different addressee labels. Better view in color.}
    \label{fig:case2-2}
\end{figure}

Figure \ref{fig:case2-2} presents another example sampled from the testing set, where we investigate how different addressee labels affect the generated responses. In the figure, different colors represent different utterances in the \emph{Dialogue History} part, and different responses generated by giving the corresponding utterances as addressees in the \emph{Generated Responses} part. This conversation is about discussing the file system in Ubuntu that can share on a network with windows machines. When the addressee is given as $\bm{U_1}$, our model suggests using \emph{samba}, which is a solution to the question of $\bm{U_1}$. Responses to $\bm{U_2}$ and $\bm{U_3}$ are like safe responses, but they make sense in their contexts: the former expresses its confusion about a confusing utterance ($\bm{U_2}$), and the latter expresses its gratitude to the suggestion in $\bm{U_3}$. Response to $\bm{U_4}$ states his/her understanding towards $\bm{U_4}$, and questions if his/her understanding is right. Response to $\bm{U_5}$ acknowledges the solution \emph{gentoo} in $\bm{U_5}$ by saying \emph{using gentoo on my computer too}. In general, this case demonstrates the ability of our model to generate diverse responses according to the specified addressees and contexts of the dialogue history.

\subsection{Response Parser: A Byproduct for Free}
Another contribution of our EM pre-training is that a response parser can be freely obtained. This byproduct comes from Eq. (\ref{eq:e_step}), where given a response generation model with addressee modeling, we can predict the addressee for each utterance in the dialogue. Previous literature has studied and proved that explicitly modeling the structural information is beneficial to understanding specific structured data. \cite{li-etal-2020-molweni, li-etal-2022-semantic, li-etal-2022-back}. In this context, the response parser can be used to infer the discourse structures, which contributes to boosting the performance of some multi-party dialogue comprehension tasks like response selection and question answering. \cite{thread-kenny, li-zhao-2021-self-pseudo, maxb-disentangle}

\section{Conclusion}
Most multi-party dialogue corpora are not annotated with addressee labels, making them unable to support the pre-training of response generation models. To solve this problem, we design a simple yet effective way to model the addressee of a response as a latent variable and propose an EM pre-training approach that iteratively performs the expectation steps to generate addressee labels, and the maximization steps to optimize a response generation model. Mathematical derivation, experimental results on the Ubuntu IRC benchmark, and extensive analyses have justified the theoretical feasibility and actual effectiveness of our method.

\section*{Limitations}
First, Due to the lack of datasets to evaluate the MPDRG task, we perform our experiments only on the Ubuntu IRC benchmark and pre-train our model only on the domain of Ubuntu chats. However, the potential of our approach goes far beyond that since it is applicable to any open-domain multi-party dialogue dataset. In the future work, we will consider applying our method in more open-domain conversational datasets, such as the transcripts of TV series or movies.

Additionally, the pre-training process solely relies on the addressee information of individual turns, disregarding the reply-to relations within the dialogue history. This oversight prevents the model from benefiting from valuable contextual cues necessary for a comprehensive understanding of the multi-party dialogue. In our future work, we will explore the integration of discourse-level reply-to relations into the pre-training process to further enrich the capabilities of the model.

\bibliography{anthology,custom}
\bibliographystyle{acl_natbib}

\end{document}